\documentclass[sigconf,natbib=true]{acmart}

\copyrightyear{2026}
\acmYear{2026}
\setcopyright{cc}
\setcctype{by}
\acmConference[CHIIR '26]{2026 ACM SIGIR Conference on Human Information Interaction and Retrieval}{March 22--26, 2026}{Seattle, WA, USA}
\acmBooktitle{2026 ACM SIGIR Conference on Human Information Interaction and Retrieval (CHIIR '26), March 22--26, 2026, Seattle, WA, USA}
\acmPrice{}
\acmDOI{10.1145/3786304.3787923}
\acmISBN{979-8-4007-2414-5/2026/03}

\begin{document}

\title[Supporting Humans in Evaluating AI Summaries of Legal Depositions]{Supporting Humans in Evaluating AI Summaries \\ of Legal Depositions}

\author{Naghmeh Farzi}
\email{Naghmeh.Farzi@unh.edu}
\orcid{https://orcid.org/0009-0000-3297-8888}
\affiliation{%
  \institution{University of New Hampshire}
  \city{Durham}
  \state{New Hampshire}
  \country{USA}
}
\author{Laura Dietz}
\email{dietz@cs.unh.edu}
\orcid{https://orcid.org/0000-0003-1624-3907}
\affiliation{%
  \institution{University of New Hampshire}
  \city{Durham}
  \state{New Hampshire}
  \country{USA}
}
\author{Dave D. Lewis}
\email{dlewis@nextpoint.com}
\orcid{https://orcid.org/0009-0000-6656-8510}
\affiliation{%
  \institution{Nextpoint}
  \city{Chicago}
  \state{Illinois}
  \country{USA}
}


\begin{abstract}
While large language models (LLMs) are increasingly used to summarize long documents, this trend poses significant challenges in the legal domain, where the factual accuracy of deposition summaries is crucial. Nugget‑based methods have been shown to be extremely helpful for the automated evaluation of summarization approaches. In this work, we translate these methods to the user side and explore how nuggets could directly assist end users. Although prior systems have demonstrated the promise of nugget‑based evaluation, its potential to support end users remains underexplored. Focusing on the legal domain, we present a prototype that leverages a factual nugget‑based approach to support legal professionals in two concrete scenarios: (1) determining which of two summaries is better, and (2) manually improving an automatically generated summary.\footnote{Prototype available at \url{https://github.com/TREMA-UNH/human-ai-legal-summaries}}
\end{abstract}


\maketitle

\section{Introduction}

Deposition transcripts are verbatim records of witness testimony taken under oath during pre‑trial discovery.  A single transcript often exceeds hundreds of pages and contains dense question and answer exchanges.  Legal professionals must manually extract, synthesize, and validate the critical facts, a process that is time‑consuming, error‑prone, and cognitively demanding.  Conflicting testimony, inconsistent statements, and the need to cross‑reference other depositions or exhibits compound the difficulty.  Cognitive fatigue and tight deadlines increase the risk that essential details are missed, creating a major efficiency bottleneck.

Large language models (LLMs) have opened new possibilities for automated text analysis, but may contain hallucinations~\cite{Xu_Jain_Kankanhalli_2025}, factual inaccuracies, and a lack of transparency, which necessitates rigorous verification~\cite{Yu_Alì_2019}.  Explainable and transparent AI is therefore essential in the legal domain.

We focus on early-stage litigation review, where AI-generated deposition summaries support case assessment. In this setting, legal professionals face two high-cognitive-load scenarios when using AI-generated summaries: 
(1) Determining which of two AI-generated summaries is better, often arising from different summary styles (e.g., chronological vs. narrative), prompting strategies, model versions, or LLM non-determinism; (2) Manually improving an AI-generated summary. Currently, both tasks are performed without system support.  The second requires exhaustive cross‑checking against the transcript, and the first involves direct comparison of the summaries together with verification against the source.

Recent advances in evaluating summarization and retrieval-augmented generation (RAG) systems focus on system‑level performance but provide a promising foundation for user‑focused assistance.  Traditional metrics such as ROUGE~\cite{rouge2004} and BLEU~\cite{Papineni_Roukos_Ward_Zhu_2002}  rely on lexical overlap, while manually aligned nuggets have been used for finer‑grained evaluation~\cite{lin2006will}. Recent automatic evaluation methods use nuggets with LLM-based alignment to evaluate the recall of facts~\cite{farzi2024exampp,Walden_Weller_Dietz_Li_Liu_Hou_Yang_2025, Pradeep_Thakur_Upadhyay_Campos_Craswell_Lin_2024, Pradeep_Thakur_Upadhyay_Campos_Craswell_Lin_2025}. Many of these newer methods employ an ``LLM-as-a-Judge'' framework, which uses LLMs as automated evaluators to assess AI responses against human-like criteria~\cite{Zheng2023judging, multicriteria}.  Although effective for model evaluation, these approaches have not yet been adapted to directly support end‑user tasks.  In this work, we leverage nugget‑based evaluation to help legal professionals verify, compare, and refine AI‑generated summaries.

\paragraph{Contributions.} 
We present a demonstration prototype that integrates automatic nugget extraction and alignment, and citation verification, are integrated into a user interface that supports legal experts in two cognitively demanding scenarios: (1) determining which of two AI summaries is better, and (2) manually improving an AI‑generated draft.

\section{Related Work}
\label{sec:relwork}
Our work sits at the intersection of legal information retrieval (IR) and extraction, and human-in-the-loop systems. We ground our contributions in a review of existing literature, identifying persistent challenges that our framework is designed to address.

\subsection{Legal Text Summarization}

Early efforts in legal summarization primarily focus on extractive methods, where key sentences are selected from a source document. Systems like CaseSummarizer~\cite{Polsley_Jhunjhunwala_Huang_2016} exemplify this approach, which uses word frequency and domain-specific rules. The advent of deep learning led to abstractive summarization models, with Norkute et al.~\cite{Norkute_Herger_Michalak_Mulder_Gao_2021} using a Pointer Generator network and Zhong et al.~\cite{Zhong_Zhong_Zhao_Wang_Ashley_Grabmair_2019} exploring an approach named 'train-attribute-mask' to identify predictive sentences. However, as they noted, focusing solely on predictive sentences may not capture all legally relevant facts, while summarizing legal meetings, depositions, and court transcripts has been largely overlooked, underscoring a gap in domain-specific applications~\cite{Akter_Çano_Weber_Dobler_Habernal_2025}.

\subsection{Hallucination and Factual Inaccuracy}
The move toward fact-based evaluation is driven by the risk of hallucination and factual inaccuracy, which is a major concern in high-stakes domains such as legal. A landmark study by Magesh et al.~\cite{Magesh_Surani_Dahl_Suzgun_Manning_Ho_2024} revealed that leading AI legal research tools, such as Casetext, Thomson Reuters, or LexisNexis, despite using advanced Retrieval-Augmented Generation (RAG) techniques, produced hallucinated content in over 17\% of the time. Similarly, Savelka et al.~\cite{Savelka_Ashley_Gray_Westermann_Xu_2023} confirmed that direct use of models like GPT-4, without retrieval-augmentation, resulted in factual inaccuracies. These findings highlight that unsupervised AI systems are not yet suitable for critical legal tasks without verification, a sentiment supported by Akter et al.~\cite{Akter_Çano_Weber_Dobler_Habernal_2025} who noted the risk of missing or misrepresenting critical facts in legal summarization. Accordingly, we treat hallucinations as inherent and focus on surfacing omissions and unsupported claims to aid human verification.

\subsection{Evaluating Legal Summaries}

A critical finding across the literature is that standard automated metrics such as ROUGE~\cite{rouge2004} and BLEU~\cite{Papineni_Roukos_Ward_Zhu_2002} are inadequate for evaluating legal summaries. These metrics offer little explainability beyond surface-level word matching and fail to capture paraphrases, synonyms, and some other critical dimensions of quality, such as contextual fidelity~\cite{Steffes_Rataj_Burger_Roth_2023}, factual accuracy, and traceability to the source text. This has spurred a move toward fact-based evaluation frameworks. Some of the recent work relies on using questions and answers to do the evaluation task. Farzi and Dietz~\cite{Farzi_Dietz_2024} use questions as a rubric and their answerability against the document as a way to evaluate the LLM or retriever's output. Xu and Ashley~\cite{Xu_Ashley_2023} proposed a question-answering approach to evaluate legal case summaries that uses an LLM to generate and grade question and answer pairs from summaries, relying on comparing answers derived from the human-written reference summary with answers generated from the model-generated summary to assess quality. ARGUE~\cite{Mayfield_Yang_Lawrie_MacAvaney_McNamee_Oard_Soldaini_Soboroff_Weller_Kayi_et_al._2024} and its automated implementation Auto-ARGUE~\cite{Walden_Weller_Dietz_Li_Liu_Hou_Yang_2025}, while not specifically for legal domain evaluation, focus on metrics such as sentence precision and nugget recall, by an LLM prompted to make binary judgments on whether a given fact is present or a claim is supported by its citation. While prior work uses fact-based rubrics to generate static evaluation scores (e.g., nugget recall), our system uses its nugget-based rubric to enable two user-centered workflows to support them for summary comparison and refinement.

\subsection{Human-in-the-Loop (HITL) for Verification}

The inherent unreliability of current AI systems underscores the critical need for human oversight. Our work builds on this insight by integrating human oversight directly into the evaluation process. Prior literature also supports this: Norkute et al.~\cite{Norkute_Herger_Michalak_Mulder_Gao_2021}, for example, demonstrated that explainability features such as highlighting source text significantly improved user trust and reduced editing time for legal experts.

However, passive explainability, to show a user a potential error (like highlighting), but does not help them fix it, does not solve the active cognitive challenges facing a user. Simply identifying a potential flaw still forces the user to manually find the correction and perform the edit, a gap that researchers identify as the difference between explainability and true actionability~\cite{Linkov_Galaitsi_Trump_Keisler_Kott_2020}. A professional is still left with two critical, high-load tasks that represent the entire human verification bottleneck: (1) Comparison, the decision-making step of selecting the better generated summary, and (2) Refinement, the correction step to make the summary factually trustworthy for real-world use.

Our system addresses the first challenge by aligning nuggets between summaries and the deposition, automating the tedious comparison (Workflow 1). Our refinement workflow (Workflow 2) addresses the second challenge by suggesting refinement suggestions based on nuggets and citation evaluations. By supporting both key decision-making processes, our framework aims to provide a more practical HITL solution.

\begin{figure*}

\includegraphics[width=0.75\linewidth]{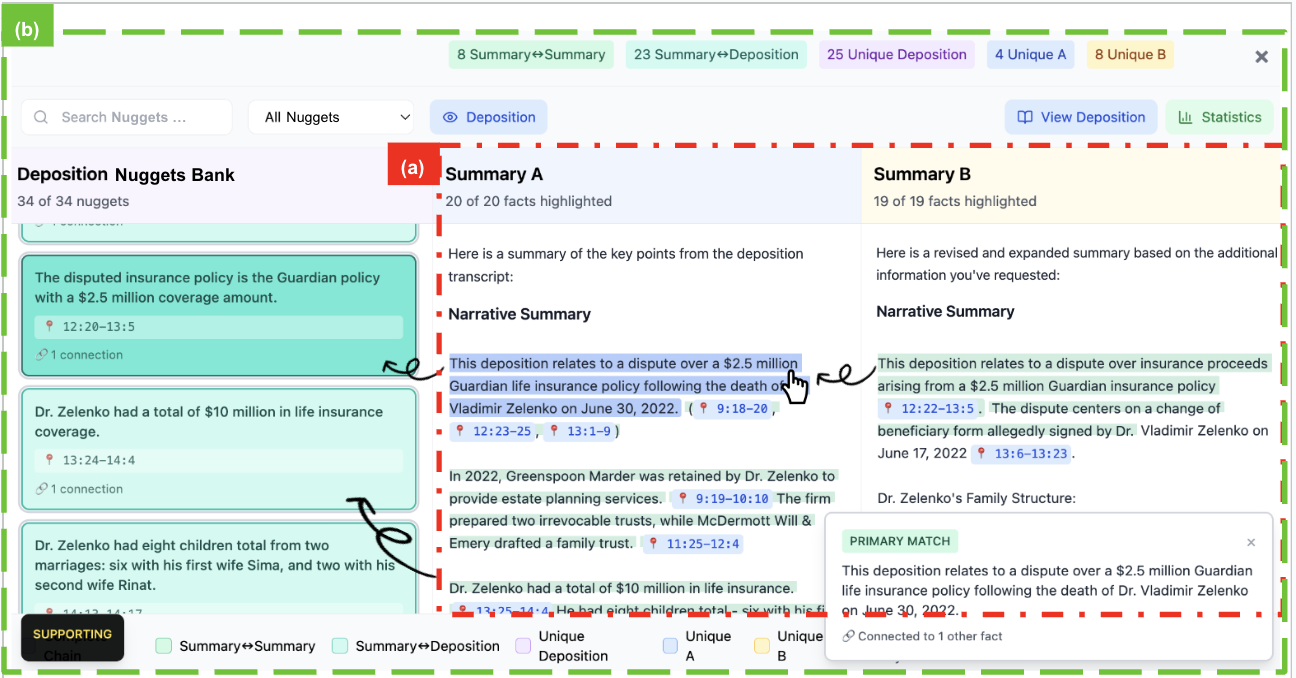}
\caption{Currently, users are left unsupported in assessing which of two summaries is better, relying only on the plain text shown in (a) without any color-coded highlights. (b) Guided Comparison interface (Scenario 1) aligns and highlights shared, unique, and missing nuggets.}
\label{fig:workflow1_compare}
\end{figure*}

\section{System Description}
\label{sec:system}

We first elaborate on how to extract and align nuggets and determine citation support in an automated way. These form the backbone on which our prototype supports two usage scenarios.

\subsection{Automated Analysis } 
Both of our demonstrated workflows are based on a shared backend pipeline designed to create a structured, verifiable rubric of nuggets from raw transcripts. This pipeline consists of two stages:

\textbf{Extracting Free-Text Nuggets:} Our prototype leverages an automatically generated nugget bank. An LLM, guided by a prompt tailored for extracting key facts from legal depositions, extracts all self-contained nugget sentences from the parsed transcript. Each extraction is accompanied by a citation specifying the supporting span of deposition with starting and ending page and line numbers. The full set forms the \emph{rubric} of nuggets.

 \textbf{Nugget alignment:}  Using an LLM-as-Judge framework, the system evaluates AI-generated summaries against the deposition nugget bank across three key criteria: completeness, citation quality, and factual accuracy.

For completeness, the system identifies which nuggets are fully or partially covered in the summary and provides detailed explanations, including specific portions of the summary that correspond to each nugget and the parts of the nugget that are absent. 

Our prototype leverages a three-point scale: fully present (2), partially present (1), or missing (0). Our prototype makes the list of partial/missing nuggets explicit, so that the user can choose which nuggets to expand to improve comprehensiveness, enabling users to address factual omissions effectively.

 \textbf{Citation support:} For citation support, the system evaluates each citation based on three metrics: accuracy (``accurate'' or ``not accurate''), coverage (``covered'' or ``not covered''), and sufficiency (``sufficient'' or ``insufficient'').

\subsection{Scenario 1: Guided Summary Comparison}

\paragraph{User's problem:} A legal professional receives two competing AI generated summaries and must decide which one to adopt. For this task the user must identify the key differences between the summaries.  

\paragraph{The challenge:}
Existing systems do not provide dedicated support for this cognitively demanding task, especially when the same facts appear in different orders or are expressed with different wording. Consequently, the user must manually read, compare, and cross-reference two dense blocks of text with the long deposition. An example is shown in Figure~\ref{fig:workflow1_compare} part~(a).  

\paragraph{How a system could help:}
By highlighting which sentences in either summary discuss the same nugget, the user can focus on a fine-grained assessment of which summary elaborates the nugget in the most useful way. 
Moreover, a nugget-based analysis can reveal discrepancies, such as a nugget that appears in one summary but is omitted by the competing summary. Together, this form of nugget-guided support will help the user manage the different pieces of information presented in both summaries.

\paragraph{Example summaries:}
Figure~\ref{fig:workflow1_compare}~(b) presents two AI‑generated summaries of the same deposition. Summary~A is produced with a prompt engineered for high conciseness and is intended for a quick overview. Summary~B is produced with a prompt engineered for high comprehensiveness and is intended for detailed case review.

\paragraph{Prototype walkthrough:}

A screenshot of our prototype is shown in Figure~\ref{fig:workflow1_compare}.
The interface presents the two summaries side by side, aligning nuggets as follows:

\begin{itemize}
\item \textbf{Matched nuggets:} nuggets that appear in both Summary~A and Summary~B. These are highlighted in green.
\item \textbf{Unique nuggets:}  nuggets that appear in only one summary. These are highlighted in blue for Summary~A and in yellow for Summary~B, emphasizing the differences.
\item \textbf{Missing nuggets:}  nuggets that exist in the deposition's nugget bank but are absent from both summaries. These are shown as potential additions to improve completeness.
\end{itemize}

By highlighting matched nuggets we enable users to easily identify overlap between summaries, while judging when nuggets are better/worse elaborated.  Unique nuggets are made salient, allowing users to assess the value of information that appears in only one summary. Missing nuggets are presented to guide users toward possible enhancements.

\begin{figure*}
    \centering
    \includegraphics[width=0.70\linewidth]{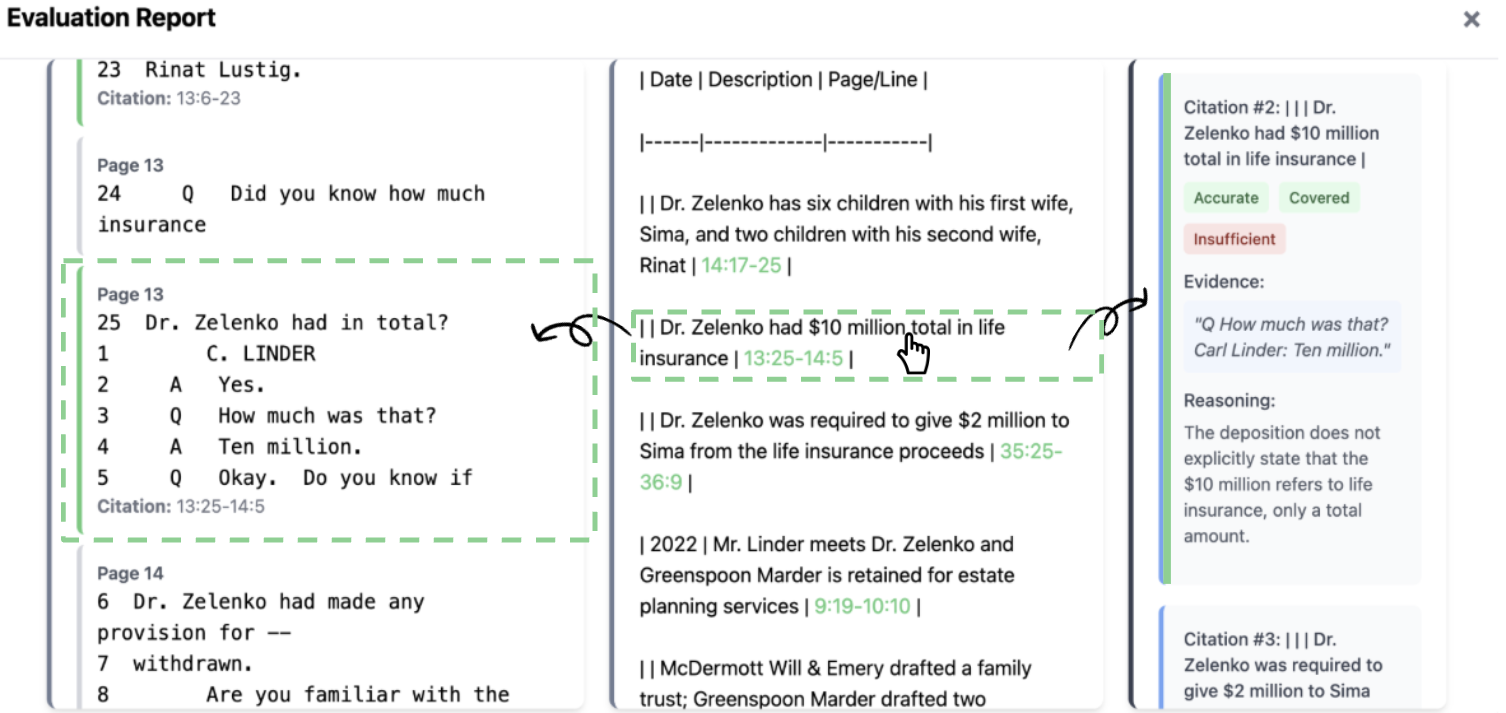}
    \caption{Middle: AI-generated summary; Left: deposition text with supporting spans highlighted; Right: evaluation results providing guidance for summary refinement.}
    \label{fig:citation_example}
\end{figure*}

The nugget bank is displayed on the left; each nugget includes its source citation. Clicking a citation jumps directly to the supporting span in the deposition. A legend at the bottom explains the color coding, and statistics at the top provide a quantitative overview of the comparison. The interface is interactive: hovering over or clicking a nugget in either summary automatically scrolls to the corresponding entry in the nugget bank and highlights the matching nugget in the other summary, offering immediate guidance for verification.

\paragraph{The vision:}
Our prototype demonstrates the value of displaying nugget annotations. We envision a deeper analysis that incorporates LLM predictions of how well a nugget is elaborated and enables the user to manually label nuggets as vital, partial, or non‑relevant.   

We also envision the extraction of hierarchical nuggets that progressively refine broader key facts into finer elements, with the system automatically selecting the granularity that most clearly reveals qualitative differences.

\subsection{Scenario 2: Guided Summary Refinement} 
\paragraph{User's problem:} A legal professional receives a single AI-generated summary that must be verified and possibly edited by adding, removing, or correcting text and citations, before it can be used.

\paragraph{The challenge:}
The user needs to efficiently locate omissions or correct the inaccuracies, but this is difficult without a comprehensive list of all relevant facts.  
 
\paragraph{How a system could help:}
By linking summary statements to their supporting nuggets and citations, the system can highlight omissions, misrepresentations, or insufficiently supported claims. This guides the user to refine both content and citations efficiently, while keeping full control over edits.

\paragraph{Prototype Walkthrough:}

The prototype first computes (1) how well each nugget is represented in the summary and (2) the degree to which citations support the summary statements. The interface conveys these results by highlighting segments that lack evidential support or contain inaccurate citations.

Figure~\ref{fig:citation_example} illustrates a case where a summary claims that the ``life insurance policy was valued at \$10 million''. However,  the deposition only supports that \$10 million amount, but does not specify ``life insurance,''. Hence, the citation is marked as insufficient, but since it accurately represents the claim of the summary, it is annotated as covered and accurate. 

For factual accuracy, the system displays discrepancies between the summary and the nugget bank and recommends double‑checking segments with potential errors. These annotations are produced through LLM‑based comparisons, providing nuanced yet automated analysis. The user is presented only with the suggested options and remains responsible for all edits.

\paragraph{The vision:} 
Users will be able to add nuggets manually to the nugget bank, after which the system will automatically highlight the relevant passages. Users may also designate nuggets as vital, thereby creating a structured to‑do list for the summary. Additional quality criteria, such as required ordering of nugget elaboration for coherence, could be captured automatically and communicated to the user, further supporting goal‑driven refinement.

\subsection{Prototype Configuration}

The prototype is preconfigured to use a  dataset comprised of nine publicly available civil litigation deposition transcripts, each averaging approximately 3,760 lines. While any LLM model can be used, our prototype uses \texttt{Claude-3.5-Sonnet}\footnote{https://www.anthropic.com/news/claude-3-5-sonnet} on Amazon Web Services (AWS) Bedrock to generate transcript summaries. Additionally, we employ  \texttt{Claude-3-Haiku}\footnote{https://www.anthropic.com/news/claude-3-haiku} with Bedrock's tool-use capabilities\footnote{https://docs.aws.amazon.com/bedrock/latest/userguide/tool-use-inference-call.html} to produce structured evaluations and consequently guidance to the end user.

\section{Conclusion}
\label{sec:conclusion}
In this paper, we sketch a vision of how automated nugget extraction can be integrated into a human-in-the-loop system to support an end user
in two critical tasks: comparing alternative AI summaries (Workflow 1) and refining a single summary (Workflow 2).  In both cases, our aim is to highlight differences and make actionable suggestions to enable the user in their workflow.

This nugget-based method, while explored for legal summaries, readily extends to other domains where users must verify, compare, or refine AI-generated content.
Our long-term vision is to integrate this structured feedback into the LLM responses, leading to better first-draft responses that are already aligned with a user's preferences.

\begin{acks}
This work is the product of a research collaboration between the University of New Hampshire and Nextpoint and was supported by a gift from Nextpoint.
\end{acks}

\bibliographystyle{ACM-Reference-Format}
\balance
\bibliography{bib}
\balance

\end{document}